\documentclass[11pt]{article}
\usepackage{coling2020}
\usepackage{times}
\usepackage{url}
\usepackage{latexsym}
\usepackage{xcolor}
\usepackage{booktabs}
\usepackage{amsmath, nccmath}
\usepackage{amssymb}
\usepackage{graphicx}

\usepackage[title,toc,titletoc,page]{appendix}

\usepackage{pifont}
\usepackage{amssymb}
\newcommand{\comment}[1]{}

\title{Knowledge Graph Embeddings in Geometric Algebras}

\author{Chengjin Xu \\
  University of Bonn / Germany \\
  {\tt  	xuc@iai.uni-bonn.de} \\\And
  Mojtaba Nayyeri \\
  University of Bonn / Germany \\
  {\tt nayyeri@iai.uni-bonn.de} \\\AND
  Yung-Yu Chen \\
    University of Bonn / Germany \\
  {\tt s6ynchen@uni-bonn.de}\And
  Jens Lehmann \\
  University of Bonn / Germany \\
  Fraunhofer IAIS/ Germany\\
  {\tt jens.lehmann@iais.fraunhofer.de}\\}
  
\date{}

\begin{document}
\maketitle
\begin{abstract}
Knowledge graph (KG) embedding aims at embedding entities and relations in a KG into a low dimensional latent representation space. Existing KG embedding approaches model entities and relations in a KG by utilizing real-valued , complex-valued, or hypercomplex-valued (Quaternion or Octonion) representations, all of which are subsumed into a geometric algebra. In this work, we introduce a novel geometric algebra-based KG embedding framework, GeomE, which utilizes multivector representations and the geometric product to model entities and relations. Our framework subsumes several state-of-the-art KG embedding approaches and 
is advantageous with its ability of modeling various key relation patterns, including (anti-)symmetry, inversion and composition, rich expressiveness with higher degree of freedom as well as good generalization capacity. Experimental results on multiple benchmark knowledge graphs show that the proposed approach outperforms existing state-of-the-art models for
link prediction.
\end{abstract}

\section{Introduction}
\label{intro}

%
%
\blfootnote{
    
    \hspace{-0.65cm}  
    This work is licensed under a Creative Commons 
    Attribution 4.0 International License.
    License details:
    \url{http://creativecommons.org/licenses/by/4.0/}.
}
Knowledge graphs (KGs) are directed graphs where nodes represent entities and (labeled) edges
represent the types of relationships among entities. This can be represented as a collection of triples $(h, r, t)$, each representing a relation $r$ between a "head-entity" $h$ and an "tail-entity" $t$. Some real-world knowledge graphs include Freebase~\cite{Freebase}, WordNet~\cite{WordNet}, YAGO~\cite{YAGO}, and DBpedia~\cite{Dbpedia}. 

However, most existing KGs are incomplete. The task of link prediction alleviates this drawback
by inferring missing facts based on the known facts in a KG and thus has gained growing interest. Embedding KGs into a low-dimensional space and learning latent representations of
entities and relations in KGs is an effective solution for this task. In general,
most existing KG embedding models learn to embed KGs by optimizing a scoring function which assigns higher scores to true facts than invalid ones. 

Recently, learning KG embeddings in the complex or hypercomplex spaces has been proven to be a highly effective inductive bias. ComplEx~\cite{ComplEx}, RotatE, pRotatE~\cite{RotatE}, and QuatE~\cite{QuatE} achieved the state-of-the-art results on link prediction, due to their abilities of capturing various relations (i.e., modeling symmetry and anti-symmetry). They both use the asymmetrical Hermitian product to score relational triples where the components of entity/relation embeddings are complex numbers or quaternions.

Complex numbers and quaternions can be described by the various components within a Clifford multivector~\cite{Multivector}. In other words, the geometric algebra of Clifford~\shortcite{clifford1882mathematical} provides an elegant and efficient rotation representation in terms of multivector which is more general than Hamilton~\shortcite{hamilton1844lxxviii}'s unit quaternion.

In this paper, we propose a novel KG embedding approach, GeomE, which is based on Clifford multivectors and the geometric product. Concretely, we utilize $N$ multivector embeddings of $N$ grades ($N=2,3$) to represent entity and relation. Each component of an entity/relation embedding is a multivector in a geometric algebra of $N$ grades, $\mathbb{G}^{N}$, with scalars, vectors and bivectors, as well as trivectors (for $N=3$). In terms of a triple $(h,r,t)$, we use an asymmetrical geometric product which involves the conjugation of the embedding of the tail entity to multiply the embeddings of $e_{s}$, $r$, $e_{o}$, and obtain the final score of the triple from the product embedding.

The advantages of our formulas include the following points:
\begin{itemize}
\setlength{\itemsep}{0pt}
\setlength{\parsep}{0.5pt}
\setlength{\parskip}{-2pt}
    \item Our framework GeomE subsumes ComplEx, pRotatE and QuatE. A complex number can be regarded as a scalar plus a bivector in the geometric algebra $\mathbb{G}^{2}$. A quaternion is isomorphic with a scalar plus three bivectors in the geometric algebra $\mathbb{G}^{3}$. Thus, GeomE inherits the excellent properties of pRotatE, ComplEx and QuatE and has the ability to model various relation patterns, e.g., (anti-)symmetry, inversion and composition.
    \item The geometric product units the Grassmann~\shortcite{grassmann2012lineale} and Hamilton~\shortcite{hamilton1844lxxviii} algebras into a single structure. Compared to the Hamilton operator used in QuatE, the geometric product provides a greater extent of expressiveness since it involves the operator for vectors, trivectors and n-vectors, in addition to scalars and bivectors.
    \item Our proposed approach GeomE is not just a single KG embedding model. GeomE can be generalized in the geometric algebras of different grades and is hence more flexible in the expressiveness compared to pRotatE, ComplEx and QuatE. In this paper, we propose two new KG embedding models, i.e., GeomE2D and GeomE3D, based on multivectors from $\mathbb{G}^{2}$ and $\mathbb{G}^{3}$, and test their combination model GeomE+.
    
\end{itemize}

Experimental results demonstrate that our approach achieves state-of-the-art results on four well-known KG benchmarks, i.e., WN18, FB15K, WN18RR, and FB15K-237.

\section{Related Work}
\label{Related Work}
Most KG embedding models can be classified as distance-based or semantic matching based, according to their scoring functions.

Distance-based scoring functions aim to learn embeddings by representing relations as translations from head entities to tail entities. Bordes et al.~\shortcite{TransE} proposed TransE by assuming that the added embedding of $s$ and $r$ should be close to the embedding of $o$. Since that, many variants and extensions of TransE have been proposed. For example, TransH~\cite{TransH} projects entities
and relations into a hyperplane. TransR~\cite{TransR} introduces separate projection spaces for entities
and relations. TransD~\cite{TransD} uses independent projection vectors for each
entity and relation and can reduce the amount of calculation compared to TransR. TorusE~\cite{toruse} defines embeddings and distance function in a compact Lie group, torus. The recent distance-based KG embedding models, RotatE and pRotatE~\cite{RotatE}, propose a rotation-based distance scoring functions with complex-valued embeddings. Likewise, TransComplEx~\cite{TransComplex} also maps entities and relations into a complex-valued vector space.

On the other hand, semantic matching models include RESCAL~\cite{RESCAL}, DistMult~\cite{DISTMULT}, ComplEx~\cite{ComplEx}, SimplE~\cite{simple} and QuatE~\cite{QuatE}. In RESCAL, each relation is represented with a square matrix, while DistMult replaces it with a diagonal matrix in order to reduce the complexity. SimplE is also a simple yet effective bilinear approach for knowledge graph embedding. ComplEx embeds entities and relations in a complex space and utilizes an asymmetric Hermitian product to score triples, which is immensely
helpful in modeling various relation patterns. QuatE extends ComplEx in a hypercomplex space and replaces the Hermitian product with the Hamilton product which provides a greater extent of expressiveness. In addition, neural network based KG embedding models have also been proposed, e.g., NTN~\cite{NTN}, ConvE~\cite{ConvE}, ConvKB~\cite{ConvKB} and IteractE~\cite{interacte}.

Our proposed approach, GeomE, subsumes ComplEx, pRotatE and QuatE in the geometric algebras. In addition to the inheritance of the attractive properties of these existing KG embedding models, our approach takes advantages of the multivectors, e.g., the rich geometric meanings, the excellent representation ability and the generalization ability in the geometric algebras of different grades. Due to the above merits of the geometric algebras and multivectors, they have also been widely applied in computer vision and neurocomputing~\cite{geometricApplication}.

\section{Geometric Algebra and Multivectors}
\label{GA}
Leaning on the earlier concepts of Grassmann~\shortcite{grassmann2012lineale}’s exterior algebra and Hamilton~\shortcite{hamilton1844lxxviii}’s quaternions, Clifford~\shortcite{clifford1882mathematical} intended his geometric algebra to describe the geometric properties of scalars, vectors and eventually higher dimensional objects.
In addition to the well known scalar and vector elements, there are bivectors, trivectors, n-vectors and multivectors which are higher dimensional generalisations of vectors. An $N$-dimensional vector space $\mathbb{R}^{N}$ can be embedded in a geometric algebra of $N$ grades , $\mathbb{G}^{N}$. In this section, we take $\mathbb{G}^2$ and $\mathbb{G}^3$ as examples to introduce multivectors and some corresponding operators.

\subsection{2-Grade and 3-Grade Multivectors}
\begin{figure}[h!]
\centering
\includegraphics[width=0.6\textwidth]{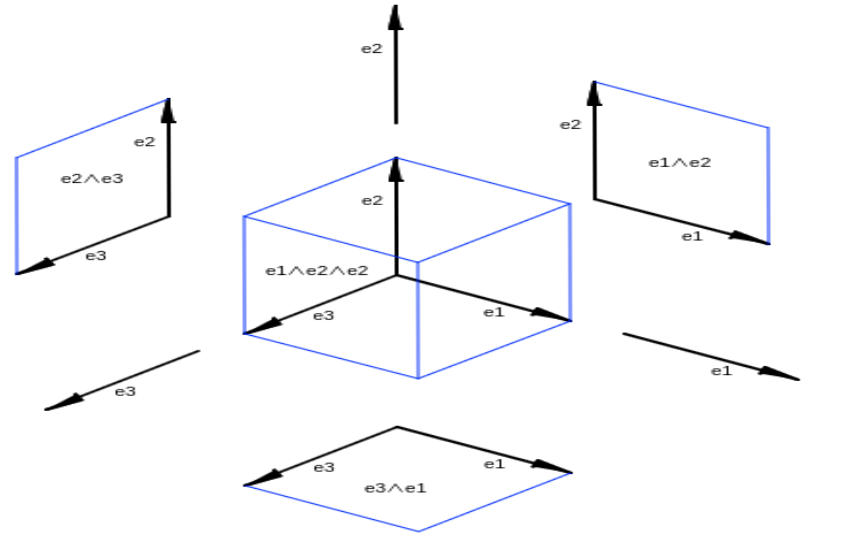} 
\caption{An example of a 3-grade multivector space $\mathbb{G}^{3}$}
\label{fig:G3}
\vspace{-0.3cm}
\end{figure}
Let $\{e_{1}, e_{2}, e_{3}\}$ be an orthonormal basis of $\mathbb{R}^{3}$. The algebra $\mathbb{G}^3$ is based on two rules: (a) $e_{i}e_{i}=1,$; (b) $e_{i}e_{j}=-e_{j}e_{i}\text{, where }i\neq j.$
The multivector space $\mathbb{G}^3$ is 8-dimensional with basis:
\vspace{-0.2cm} 
\begin{equation}
\begin{split}
    1&\text{   spans   0-vectors,   scalars,}\\
    \{e_{1},e_{2},e_{3}\}&\text{   spans   1-vectors,   vectors,}\\
    \{e_{1}e_{2},e_{2}e_{3},e_{1}e_{3}\}&\text{   spans   2-vectors,   bivectors, and}\\
    \{e_{1}e_{2}e_{3}\}&\text{   spans   3-vectors,   trivectors.}\nonumber
\end{split}
\vspace{-0.5cm} 
\end{equation}
Hence an arbitrary 3-grade multivector $M \in\mathbb{G}^3$ can be written as
\vspace{-0.2cm} 
\begin{equation}
M = a_{0}+a_{1}e_{1}+a_{2}e_{2}+a_{3}e_{3}+a_{12}e_{1}e_{2}+a_{23}e_{2}e_{3}+a_{13}e_{1}e_{3}+a_{123}e_{1}e_{2}e_{3},\nonumber
\vspace{-0.2cm} 
\end{equation}
where $a_{0},a_{1},a_{2},a_{3},a_{12},a_{23},a_{13},a_{123}$ are all real numbers.  Each element of a multivector, e.g., a scalar, a vector, or an N-vector, is called as a \textbf{blade}. A 2-grade multivector $M\in\mathbb{G}^2$ is build from one scalar, two vectors and one bivector.
\vspace{-0.2cm} 
\begin{equation}
M = a_{0}+a_{1}e_{1}+a_{2}e_{2}+a_{12}e_{1}e_{2}.\nonumber
\vspace{-0.2cm} 
\end{equation}
The norm of a multivector is equal to the root of the square sum of real values of all blades. Taking the 2-grade multivector as an example, its norm is defined as:
\vspace{-0.2cm} 
\begin{equation}
||M|| =\sqrt{a_{0}^{2}+a_{1}^{2}+a_{2}^{2}+a_{12}^{2}}.
\vspace{-0.2cm} 
\end{equation}
\subsection{Multivectors vs Quaternions}
Quaternions are elements of the form: $Q=q_{0}+q_{1}\textbf{i}+q_{2}\textbf{j}+q_{3}\textbf{k}$, where $q_0$, $q_1$, $q_2$, $q_3$ are real numbers and \textbf{i}, \textbf{j}, \textbf{k} are three different square roots of -1 and are the new elements used for the construction of quaternions. They have the following algebraic properties: $\textbf{i}^{2}=\textbf{j}^{2}=\textbf{k}^{2}=\textbf{ijk}=-1$

Bivectors from $\mathbb{G}^3$ have similar algebraic properties as the basis of the quaternion space.
\vspace{-0.2cm} 
\begin{equation}
\begin{split}
   &(e_{i}e_{j})^{2}=-e_{i}e_{j}e_{j}e_{i}=-1\text{ where } i,j = 1,2,3,\text{ and }i\neq j\\
&e_{1}e_{2}e_{2}e_{3}e_{1}e_{3}=e_{1}e_{3}e_{1}e_{3}=-1
\end{split}
\vspace{-0.5cm} 
\end{equation}

Thus we can embed a quaternion in a 3-grade geometric algebra $\mathbb{G}^3$ with a scalar plus three bivectors. A complex number can likewise be regarded as a scalar plus one bivector from $\mathbb{G}^2$.

\subsection{Geometric Product and Clifford Conjugation}
Geometric algebra also introduces a new product, \textbf{geometric product}, as well as three multivector involutions, \textbf{space inversion}, \textbf{reversion} and \textbf{Clifford conjugation}.

The geometric product of two multivectors comprises of multiplications between scalars, bivectors, trivectors and n-vectors. The product of two 2-grade multivectors $M_{a}=a_{0}+a_{1}e_{1}+a_{2}e_{2}+a_{12}e_{1}e_{2}$ and $M_{b}=b_{0}+b_{1}e_{1}+b_{2}e_{2}+b_{12}e_{1}e_{2}$ from $\mathbb{G}^2$ is equal to
\vspace{-0.2cm} 
\begin{equation}
\begin{split}
M_{a} \otimes_2 M_{b} =& a_{0}b_{0}+a_{1}b_{1}+a_{2}b_{2}-a_{12}b_{12}+(a_{0}b_{1}+a_{1}b_{0}-a_{2}b_{12}+a_{12}b_{2})e_{1}\\
&+ (a_{0}b_{2}+a_{1}b_{12}+a_{2}b_{0}-a_{12}b_{1})e_{2}+(a_{0}b_{12}+a_{1}b_{2}-a_{2}b_{1}+a_{12}b_{0})e_{1}e_{2}.
\end{split}
\vspace{-0.3cm} 
\end{equation}



The product of two 3-grade multivectors 
$M_a = a_{0}+a_{1}e_{1}+a_{2}e_{2}+a_{3}e_{3}+a_{12}e_{1}e_{2}+a_{23}e_{2}e_{3}+a_{13}e_{1}e_{3}+a_{123}e_{1}e_{2}e_{3}$
and 
$M_{b}= b_{0}+b_{1}e_{1}+b_{2}e_{2}+b_{3}e_{3}+b_{12}e_{1}e_{2}+b_{23}e_{2}e_{3}+b_{13}e_{1}e_{3}+b_{123}e_{1}e_{2}e_{3}$ from $\mathbb{G}^3$ is represented in Appendix~\ref{3-grade product}.



\textbf{Clifford Conjugation}: The Clifford Conjugation of an n-grade multivector $M$ is a subsequent composition of \textbf{space inversion} $M^*$ and \textbf{reversion} $M^{\dagger}$  as $\overline{M}=M^{\dagger*}, $
where \textbf{space inversion} $M^*$ is obtained by changing $e_i$ to $-e_i$ and \textbf{reversion} is obtained by reversing the order of all products i.e.~changing $e_{1}e_{2}\cdots e_{n}$ to $e_{n}e_{n-1}\cdots e_{1}$. 
For example, the conjugation of $M \in \mathbb{G}^{2}$, which is formed as $M=A_{0}+A_{1}+A_{2}$ with $A_{0}=a_{0}$, $A_{1}=a_{1}e_{1}+a_{2}e_{2}, A_{2}=a_{12}e_{1}e_{2}$, is computed as $\overline{M}=A_{0}-A_{1}-A_{2}$.
Note that the product of a multivector $M$ and its conjugation $\overline{M}$ is always a scalar. For a given 2-grade multivector $M = a_{0}+a_{1}e_{1}+a_{2}e_{2}+a_{12}e_{1}e_{2}$, we have
\vspace{-0.2cm} 
\begin{equation}
\begin{split}
M\otimes_2\overline {M}=a_{0}^{2}-a_{1}^{2}-a_{2}^{2}+a_{12}^{2},
\end{split}
\vspace{-0.5cm} 
\end{equation}
producing a real number, though not necessarily non-negative.




\section{Our Method}
\label{Method}
\subsection{Knowledge Graph Embedding Model based on Geometric Algebras}
Let $\mathcal{E}$ denote the set of all entities and $\mathcal{R}$ the set of
all relations present in a knowledge graph. A triple is represented as $(h, r, t)$, with $h$, $t$ $\in \mathcal{E}$ denoting head and tail entities respectively and
$r\in \mathcal{R}$ the relation between them. We use $\Omega = \{(h,r,t)\} \subseteq \mathcal{E}\times\mathcal{R}\times\mathcal{E}$ to denote the set of observed triples. 
The key issue of KG embeddings is to represent entities and relations in a continuous low-dimensional space.

Our approach GeomE uses the geometric product and multivectors for KG embedding. In this paper, we propose two models built with our approach, GeomE2D and GeomE3D, based on 2-grade multivectors and 3-grade multivectors respectively. 

\textbf{GeomE2D}  represents each entity/relation as a $k$ dimensional embedding $\textbf{M}$ where each element is a 2-grade multivector, i.e., $\textbf{M} = [M_1, \ldots, M_k], \, M_i \in{\mathbb{G}^{2}}, i=1,\ldots,k,$ where $k$ is the dimensionality of embeddings. Given a triple $(h, r, t)$, we represent embeddings of $h$, $r$ and $t$ by $\textbf{M}_{h}= [M_{h_1}, \ldots, M_{h_k}],\textbf{M}_{r}= [M_{r_1}, \ldots, M_{r_k}], \,\text{and}\, \textbf{M}_{t}= [M_{t_1}, \ldots, M_{t_k}]$ respectively.
Note that each element of $\textbf{M}$ is a 2-grade multivector. For example,  $M_{h_1} = \{h^1_{0}+h^1_{1}e_{1}+h^1_{2}e_{2}+h^1_{12}e_{1}e_{2},\, h^1_{0}, h^1_{1}, h^1_{2}, h^1_{12}\in\mathbb{
R}\}.$

\textbf{GeomE3D}  embeds $h,r,t$ into $k$ dimensional embeddings $\textbf{M}_h, \textbf{M}_r$ and $\textbf{M}_t$ respectively where each element of the embeddings is a 3-grade multivector i.e.~$M_{h_i}, M_{r_i}, M_{t_i} \in \mathbb{G}^3$ for $\, i = 1, \ldots, k,$
where $M_{h_i} = h^i_{0}+h^i_{1}e_{1}+h^i_{2}e_{2}+h^i_{3}e_{3}+h^i_{12}e_{1}e_{2}+h^i_{23}e_{2}e_{3}+h^i_{13}e_{1}e_{3}+h^i_{123}e_{1}e_{2}e_{3}.$



\textbf{Scoring Function} of GeomE is defined as the scalar of the product of the embeddings of $h$, $r$ and $t$ by using the geometric product and the Clifford conjugation.

\vspace{-0.2cm} 
\begin{equation}
\phi^{GeomE}(h,r,t)= \langle \textbf{Sc}(\textbf{M}_{h}\otimes_n \textbf{M}_{r} \otimes_n \overline{\textbf{M}_{t}}),{\textbf{1}}\rangle,
\vspace{-0.1cm} 
\label{eq:ScoreFunc}
\end{equation}
where $n = 2$ for GeomE2D and $n=3$ for GeomE3D, $\otimes_n$ denotes element-wise \textbf{Geometric Product} between two $k$ dimensional $n$-grade multivectors (e.g.~$\textbf{M}_h \otimes_n \textbf{M}_r = [M_{h_1} \otimes_n M_{r_1}, \cdots, M_{h_k} \otimes_n M_{r_k}]$), \textbf{Sc} denotes the scalar component of a multivector, ${\textbf{1}}$ denotes a $k\times1$ vector having all $k$ elements equal to one, $\overline{\textbf{M}}$ denotes the element-wise conjugation of multivectors i.e.~$\overline{\textbf{M}} = [\overline{M}_1, \ldots, \overline{M}_k]$. 
The expanded formulation of scoring functions for GeomE2D and GeormE3D are presented in the Appendix~\ref{extended scorefunction}. 

\subsection{Training}
Most of previous semantic matching models, e.g., ComplEx, are learned by minimizing a sampled binary logistic loss function~\cite{ComplEx}. Motivated by the solid results in~\cite{ComplexN3}, we formulate the link prediction task as a multiclass classification problem by using a full multiclass log-softmax loss function, and apply N3 regularization and reciprocal approaches for our models.

Given a training set $\Omega\owns(h,r,t)$, we create a reciprocal training set $\Omega^{*}\owns(t,r^{-1},h)$ by adding reverse relations and the instantaneous multiclass log-softmax loss is defined as:
\vspace{-0.2cm} 
\begin{equation}
\begin{split}
    \mathcal{L}=\sum_{(h,r,t)\in(\Omega\cup\Omega^{*})}[&-\text{log}(\frac{\text{exp}(\phi(h,r,t))}{\sum_{h^{'}\in \mathcal{E}}\text{exp}(\phi(h^{'},r,t))})-\text{log}(\frac{\text{exp}(\phi(h,r,t))}{\sum_{t^{'}\in \mathcal{E}}\text{exp}(\phi(h,r,t^{'}))})\\
    &+\frac{\lambda}{3}\sum\nolimits_{i=1}^{k}(||M_{h_{i}}||_{3}^{3}+||M_{r_{i}}||_{3}^{3}+||M_{t_{i}}||_{3}^{3})]
\end{split}
\vspace{-0.2cm} 
\end{equation}

N3 regularization and reciprocal learning approaches have been proven to be helpful in boosting the performances of semantic matching models~\cite{ComplexN3,QuatE}. Different from the sampled binary logistic loss function which generates a certain number of negative samples for each training triple by randomly corrupting the head or tail entity, the full multiclass log-softmax considers all possible negative samples and thus has a fast converge speed. On FB15K, the training process of a GeomE3D model with a high dimensionality of $k=1000$ needs less than 100 epochs and cost about 4 minutes per epoch on a single GeForce RTX 2080 device.
\subsection{Connection to QuatE, Complex and pRotatE}
As mentioned in Section~\ref{GA}, a bivector unit in a geometric algebra has similar properties to an imaginary unit in a complex or hypercomplex space. Thus, a quaternion is isomorphic with a 3-grade multivector consisting of a scalar and three bivectors, and a complex value can be regarded as a 2-grade multivector consisting of a scalar and one bivector.

\textbf{Subsumption of QuatE}:
By setting the coefficients of vectors and trivectors of $\mathbf{M}_{h}$, $\mathbf{M}_{r}$, and $\mathbf{M}_{t}$ in Equation~\ref{eq:ScoreFunc} to zero, we obtain the following equations for GeomE3D

\vspace{-0.1cm} 
\begin{equation}
\scalebox{.88}{$
\begin{split}
&\phi^{GeomE3D}(h,r,t)=\\
&(\mathbf{h}_{0}\circ \mathbf{r}_{0}-\mathbf{h}_{12}\circ \mathbf{r}_{12}-\mathbf{h}_{23}\circ \mathbf{r}_{23}-\mathbf{h}_{13}\circ \mathbf{r}_{13})\circ \mathbf{t}_{0}+(\mathbf{h}_{0}\circ \mathbf{r}_{12}+\mathbf{h}_{12}\circ \mathbf{r}_{0}-\mathbf{h}_{13}\circ \mathbf{r}_{23}+\mathbf{h}_{23}\circ \mathbf{r}_{13})\circ \mathbf{t}_{12}+\\
&(\mathbf{h}_{0}\circ \mathbf{r}_{23}+\mathbf{h}_{23}\circ \mathbf{r}_{0}-\mathbf{h}_{12}\circ \mathbf{r}_{13}+\mathbf{h}_{13}\circ \mathbf{r}_{12})\circ \mathbf{t}_{23}+(\mathbf{h}_{0}\circ \mathbf{r}_{13}+\mathbf{h}_{13}\circ \mathbf{r}_{0}+\mathbf{h}_{12}\circ \mathbf{r}_{23}-\mathbf{h}_{23}\circ \mathbf{r}_{12})\circ \mathbf{t}_{13}.
\end{split}$
}
\label{QuatE}
\end{equation}
where $\circ$ denotes Hadamard product,  $\mathbf{h}_j = [h^1_j, \ldots, h^k_j], \,\, j \in \{0, 12, 23, 13\}$.
We can find that Equation~\ref{QuatE} recovers the form of the scoring function of QuatE regardless of the normalization of the relational quaternion. Therefore, GeomE3D \emph{subsumes} the QuatE model.

\textbf{Subsumption of ComplEx}:
By setting the coefficient of vectors $\mathbf{M}_{h}$, $\mathbf{M}_{r}$, and $\mathbf{M}_{t}$ to zero in Equation~\ref{eq:ScoreFunc} for GeomE2D, we obtain

\vspace{-0.2cm} 
\begin{equation}
\scalebox{.92}{$
\begin{split}
\phi^{GeomE2D}(h,r,t)= (\mathbf{h}_{0}\circ \mathbf{r}_{0}-\mathbf{h}_{12}\circ \mathbf{r}_{12})\circ \mathbf{t}_{0}+(\mathbf{h}_{0}\circ \mathbf{r}_{12} + \mathbf{h}_{12}\circ \mathbf{r}_{0})\circ \mathbf{t}_{12},
\end{split}
$}
\vspace{-0.3cm}
\label{Complex}
\end{equation}
where $\mathbf{h}_j = [h^1_j, \ldots, h^k_j], \,\, j \in \{0, 12\}$.
The Equation~\ref{Complex} recovers the form of the scoring function of ComplEx. Therefore, GeomE2D \emph{subsumes} the ComplEx model.
Additinally, comparing equations~\ref{QuatE} and \ref{Complex}, we conclude that GeomE3D also \emph{subsumes} ComplEx.

\textbf{Subsumption of pRotatE}:
Apart from ComplEx and QuatE, GeomE also subsumes pRotatE. We start from the formulation of the scoring function of pRotatE and show that the scoring function is a special case of Equation~ \ref{Complex}. The scoring function of pRotatE is defined as
\begin{equation}
\begin{split}
    \phi^{pRotatE}(h,r,t) = -\| \mathbf{h}\circ \mathbf{r} - \mathbf{t}\|,
    \end{split}    \label{pRotatE}
\end{equation}
where the modulus of each element of relation vectors is $|\mathbf{r}_i| = 1, i=1,\dots,k,$ and $|\mathbf{h}_i| = |\mathbf{t}_i| = C \in \mathbb{R}^+.$
After some derivation on the score of pRotatE and GeomE2D (see details in Appendix~\ref{proof pRotatE}), we can obtain $\phi^{GeomE2D}(h,r,t) = \frac{\phi^{pRotatE^2}(h,r,t)-2kC^2}{2}$.
Note that $2kC^2$ is a constant number as $k$ and $C$, and thus does not affect the overall ranking obtained by computing and sorting the scores of triples. For a triple $(h,r,t)$, there is a positive correlation between its GeomE score and pRotatE score since pRotatE scores are always non-positive. Therefore, GoemE2D and consequently GeomE3D \emph{subsumes} the pRotatE model in the terms of ranking.




Overall, it can be seen that our framework subsumes ComplEx, pRotatE and QuatE and provides more degrees of freedom by introducing vectors and trivectors.
In addition, our framework can be generalized into the geometric algebras with higher grades ($\mathbb{G}^n, n > 3$) and is hence more flexible in expressiveness.

Although we introduce more coefficients in our framework, our models have the same time complexity as pRotatE, ComplEx and QuatE as shown in Table~\ref{complexity}. And the memory sizes of our models increase linearly with the dimensionality of embeddings.

\begin{table}[h]
\begin{center}
\resizebox{0.8\textwidth}{!}{
\begin{tabular}{ccccc}
\hline  Model & Scoreing Function & Relation Parameters &  $\mathcal{O}_{time}$&$\mathcal{O}_{space}$\\ \hline
TransE&$-||\mathbf{h+r-t}||$&$\mathbf{r}\in \mathbb{R}^{k}$&$\mathcal{O}(k)$&$\mathcal{O}(k)$\\
DistMult&$<\mathbf{h,r,t}>$&$\mathbf{r}\in \mathbb{R}^{k}$&$\mathcal{O}(k)$&$\mathcal{O}(k)$\\
ComplEx&\textbf{Re}$(\mathbf{<h,r,\overline{t}>})$&$\mathbf{r}\in \mathbb{C}^{k}$&$\mathcal{O}(k)$&$\mathcal{O}(k)$\\
ConvE&$f(\text{vec}(f([\mathbf{W_{h};W_{r}}]\ast\boldsymbol{\mathcal{\omega}}))\textbf{W})\mathbf{t}$&$\mathbf{W_{r}}\in\mathbb{R}^{k_{w}\times k_{h}}$&$\mathcal{O}(k)$&$\mathcal{O}(k)$\\
(p)RotatE&$-||\mathbf{h\circ r-t}||$&$\mathbf{r}\in \mathbb{C}^{k}$&$\mathcal{O}(k)$&$\mathcal{O}(k)$\\\
QuatE&$\mathbf{Q_{h}\otimes W_{r}^{\triangleleft}\cdot Q_{t}}$&$\mathbf{W_{r}}\in\mathbb{H}^{k}$&$\mathcal{O}(k)$&$\mathcal{O}(k)$\\
\hline
\specialrule{0em}{1pt}{1pt}
GeomE2D&$\langle \textbf{Sc}(\textbf{M}_{h}\otimes_2 \textbf{M}_{r} \otimes_2 \overline{\textbf{M}_{t}}),{\textbf{1}}\rangle$&$\mathbf{M_{r}}\in\mathbb{G}^{2\times k}$&$\mathcal{O}(k)$&$\mathcal{O}(k)$\\
GeomE3D&$\langle \textbf{Sc}(\textbf{M}_{h}\otimes_3 \textbf{M}_{r} \otimes_3 \overline{\textbf{M}_{t}}),{\textbf{1}}\rangle$&$\mathbf{M_{r}}\in\mathbb{G}^{3\times k}$&$\mathcal{O}(k)$&$\mathcal{O}(k)$\\

\hline
\end{tabular}}
\end{center}
\caption{Scoring functions of state-of-the-art link prediction models, their parameters as well as their time complexity and space complexity. $\text{vec}()$ denotes the matrix flattening. $\ast$ denotes the convolution operator. $f$ denotes a non-linear function. $\otimes$ denotes the Hamilton product. $\mathbb{H}$ denotes a hypercomplex space. ${\triangleleft}$ denotes the normalization of quaternions.}\label{complexity}
\vspace{-0.3cm}
\end{table}
\subsection{Ability of Modeling Various Relation Patterns}
\label{relation_modeling}
Our framework subsumes pRotatE, ComplEx and QuatE, and thus inherits their attractive properties: 
One of the merits of our framework is the ability of modeling various patterns including \textbf{symmetry/anti-symmetry, inverse} and \textbf{composition}. 
We give the formal definitions of these relation patterns.\\
\textbf{Definition  } \textit{A relation r is \textbf{symmetric (anti-symmetric)}, if $\forall h,t\ \  r(h,t)\Rightarrow r(t,h)\  (r(h,t)\Rightarrow \neg r(t,h))$.}\\
\textbf{Definition  } \textit{A relation $r_1$ is \textbf{inverse} of relation $r_2$, if $\forall h,t\ \  r_1(h,t)\Rightarrow r_2(t,h)$.\\}
\textbf{Definition  } \textit{relation $r_3$ is \textbf{composed of} relation $r_1,r_2$, if $\forall h,o,t\ \  r_1(h,o) \land r_2(o, t) \Rightarrow r_3(h, t)$.\\
Clauses with the above-mentioned forms are  \textbf{(anti-)symmetry, inversion} and \textbf{composition} patterns.}\\


GeomE can infer and model various relation patterns defined above by taking advantages of the flexibility and representational power of geometric algebras and the geometric product.

\textbf{(Anti-)symmetry:} By utilizing the conjugation of embeddings of tail entities, our framework can model (anti-)symmetry patterns. The symmetry property of GeomE2D and GeomE3D can be proved by enforcing the coefficients of vectors and bivectors in relation embeddings to be zero. On the other hand, their scoring functions are asymmetric about relations when the coefficients of vectors and bivetors in relation embeddings are nonzero. For GeomE2D, the difference score of $(h,r,t)$ and $(t,r,h)$ is equal to
\begin{equation}
\scalebox{.95}{$
\begin{split}
&\phi^{GeomE2D}(h,r,t)-\phi^{GeomE2D}(t,r,h)= 2[(\mathbf{h}_{1}\circ \mathbf{t}_{0}-\mathbf{h}_{0}\circ \mathbf{t}_{1}+\mathbf{h}_{12}\circ \mathbf{t}_{2}-\mathbf{h}_{2}\circ \mathbf{t}_{12})\circ \mathbf{r}_{1}+\\
&(\mathbf{h}_{2}\circ \mathbf{t}_{0}-\mathbf{h}_{0}\circ \mathbf{t}_{2}+\mathbf{h}_{1}\circ \mathbf{t}_{12}-\mathbf{h}_{12}\circ \mathbf{t}_{1})\circ \mathbf{r}_{2}+(\mathbf{h}_{0}\circ \mathbf{t}_{12}-\mathbf{h}_{12}\circ \mathbf{t}_{0}+\mathbf{h}_{2}\circ \mathbf{t}_{1}-\mathbf{h}_{1}\circ \mathbf{t}_{2})\circ \mathbf{r}_{12}].
\end{split}$
}
\label{symmetry}
\end{equation}
This difference is equal to zero when $\mathbf{r}_{1},\mathbf{r}_{2},\mathbf{r}_{12}=\textbf{0}$. Embeddings of multiple symmetric relations could still express their different semantics since their scalar parts might be different.

\textbf{Inversion:} As for a pair of inverse relations $r$ and $r'$, the scores of $(h,r,t)$ and $(t,r',h)$ are equal 
when $\mathbf{M}_{r}$= $\overline{\mathbf{M}_{r'}}$. 
Concretely, the difference score of $(h,r,t)$ and $(t,r',h)$ is equal to
\begin{equation}
\scalebox{.95}{$
\begin{split}
&\phi^{GeomE2D}(h,r,t)-\phi^{GeomE2D}(t,r',h)= (\mathbf{h}_{1}\circ \mathbf{t}_{0}-\mathbf{h}_{0}\circ \mathbf{t}_{1}+\mathbf{h}_{12}\circ \mathbf{t}_{2}-\mathbf{h}_{2}\circ \mathbf{t}_{12})\circ (\mathbf{r}_{1}+\mathbf{r'}_{1})+\\
&(\mathbf{h}_{2}\circ \mathbf{t}_{0}-\mathbf{h}_{0}\circ \mathbf{t}_{2}+\mathbf{h}_{1}\circ \mathbf{t}_{12}-\mathbf{h}_{12}\circ \mathbf{t}_{1})\circ (\mathbf{r}_{2}+\mathbf{r'}_{2})+(\mathbf{h}_{0}\circ \mathbf{t}_{12}-\mathbf{h}_{12}\circ \mathbf{t}_{0}+\mathbf{h}_{2}\circ \mathbf{t}_{1}-\mathbf{h}_{1}\circ \mathbf{t}_{2})\\
&\circ (\mathbf{r}_{12}+\mathbf{r}'_{12})+(\mathbf{h}_{0}\circ \mathbf{t}_{0}-\mathbf{h}_{1}\circ \mathbf{t}_{1}-\mathbf{h}_{2}\circ \mathbf{t}_{2}+\mathbf{h}_{12}\circ \mathbf{t}_{12})\circ(\mathbf{r}_{0}-\mathbf{r'}_{0}).
\end{split}$
}
\label{inverse}
\end{equation}
This difference is equal to zero when $\mathbf{r}_{0}=\mathbf{r'}_{0}, \mathbf{r}_{1}=-\mathbf{r'}_{1},\mathbf{r}_{2}=-\mathbf{r'}_{2},\mathbf{r}_{12}=-\mathbf{r'}_{12}$.

\textbf{Composition:} GeomE can also model composition patterns by introducing some constraints on embeddings. The detailed proof can be found in Appendix~\ref{proof in Matrix}.
\section{Experiments and Results}
\subsection{Experimental Setup}
\paragraph{Datasets} We use four widely used KG benchmarks for evaluating our proposed models, i.e., FB15K, WN18, FB15K-237 and WN18RR. The statics of these datasets are listed in Table~\ref{statics}. FB15K and WN18 are introduced in~\cite{TransE}. The former is extracted from FreeBase~\cite{Freebase}, and the latter is a subsampling of WordNet~\cite{WordNet}. It is firstly discussed in~\cite{FB15K237}
that WN18 and FB15K suffer from test leakage through inverse relations, i.e. many test triples can be obtained simply
by inverting triples in the training set. To address this issue,
Toutanova et al.~\shortcite{FB15K237} generated FB15K-237 by removing inverse relations in FB15K. Likewise, Dettmers et al~\cite{ConvE} generated WN18RR by removing inverse relations in WN18. The recent literature shows that FB15K-237 and WN18RR are harder to fit and thus more challenging for new KG embedding models.
The details of dataset statistics are listed in the Appendix~\ref{datasets}.
\paragraph{Evaluation Protocols} Link prediction is to complete a fact with a missing entity. Given a test triple $(h, r, t)$, we corrupt this triple by replacing $h$ or $t$ with all possible entities, sort all the corrupted triples based on their scores and compute the rank of the test triple. Three evaluation metrics are used here, \textbf{Mean Rank (MR)}, \textbf{Mean Reciprocal Rank (MRR)} and \textbf{Hits@k}. We also apply the filtered setting proposed in~\cite{TransE}.

\paragraph{Implementation Details} We used Adagrad~\cite{Adagrad} as the optimizer and fine-tuned the
hyperparameters on the validation dataset. We fixed batch size $b=1000$ and learning rate $lr=0.1$. We decided to focus on influences of the embedding dimensionality $k\in\{20,50,100,200,500,1000\}$ and the regularization coefficient $\lambda \in \{0.0001,0.00025,0.0005,0.00075,0.001,\cdots,0.1\}$. The default configuration for our proposed models is as  follows: $lr=0.1$, $b=1000$, $k=1000$ and $\lambda=0.01$. Below, we only list the non-default hyperparameters for both GeomE2D and GeomE3D: $\lambda=0.025$ on WN18, $\lambda=0.05$ on FB15K-237, and $\lambda=0.1$ on WN18RR. We implemented our model using PyTorch~\cite{pytorch} and ran the training processes on a single GeForce RTX 2080 GPU. To prevent over-fitting, we used the early-stop setting on validation set and set the maximum epoch to 100.

\paragraph{Baselines}We compare our models against a variety of baselines including:   DistMult~\cite{DISTMULT}, ComplEx~\cite{ComplEx}, R-GCN+~\cite{RGCN}, ConvE~\cite{ConvE}, SimplE~\cite{simple}, TorusE~\cite{toruse}, RotatE, pRotatE~\cite{RotatE}, InteractE~\cite{interacte} and QuatE$^{2}$~\cite{QuatE}. We choose QuatE$^{2}$ as baseline since this variant of QuatE applies N3 regularization and reciprocal approaches as our models and get the best results among all variants of QuatE. Lacroix et.al.,~\shortcite{ComplexN3} also uses these approaches to boost the performance of ComplEx. The results reported in~\cite{ComplexN3} are quite close to QuatE$^2$ regarding MRR and Hits@10. Apart from GeomE2D and GeomE3D, we test a combination model GeomE+ by utilizing the ensemble method used for DistMult~\cite{DistMultResult} and R-GCN+~\cite{RGCN} to ensemble GeomE2D and GeomE3D. A GeomE+ model consists of a GeomE2D model plus a GeomE3D model which are separately trained with the optimal hyperparameters, i.e., $\phi^{GeomE+}(h,r,t)=\phi^{GeomE2D}(h,r,t)+\phi^{GeomE3D}(h,r,t)$.

\subsection{Experimental Results}

\begin{table}[h]
\begin{center}
\resizebox{\textwidth}{!}{
\begin{tabular}{ccccccccccc}
\hline
 &\multicolumn{5}{c}{FB15K}&\multicolumn{5}{c}{WN18}\cr 
 \cmidrule(lr){2-6}\cmidrule(lr){7-11}
&MR&MRR&Hits@1&Hits@3&Hits@10& MR&MRR&Hits@1&Hits@3&Hits@10\\
\hline
DistMult*&42& 0.798 &- &- &0.893 &655 &0.797 &- &- &0.946\\
ComplEx&-& 0.692& 0.599& 0.759&0.840&-&0.941&0.930&0.945&0.947\\
ConvE&51&0.657&0.558&0.723&0.831&374&0.943&0.935&0.946&0.956\\
R-GCN+&-&0.696&0.601&0.760&0.842&-&0.819&0.697&0.929&\textbf{0.964}\\
SimplE&-&0.727&0.660&0.773&0.838&-&0.942& 0.939&0.944&0.947\\
TorusE&-&0.733&0.674&0.771&0.832&-&0.947&0.943&0.950&0.954\\
RotatE&40&0.797&0.746&0.830&0.884&309&0.949&0.944&0.952&0.959\\
pRotatE&43&0.799&0.750&0.829&0.884&\textbf{254}&0.947&0.942&0.950&0.957\\
QuatE$^{2}$&-&0.833&0.800&0.859&0.900&-&0.950&0.944&0.954&0.962\\
\hline
GeomE2D&34&0.853&0.816&0.877&0.913&259&0.951&0.946&0.954&0.960\\
GeomE3D&36&0.846&0.806&0.876&0.915&325&0.951&\textbf{0.947}&0.954&0.959\\
GeomE+&\textbf{30}&\textbf{0.854}&\textbf{0.817}&\textbf{0.880}&\textbf{0.916}&\textbf{254}&\textbf{0.952}&\textbf{0.947}&\textbf{0.955}&0.962\\
\hline
\end{tabular}
}
\end{center}
\vspace{-0.1cm}
\caption{Link prediction results on FB15K and WN18. * indicates that results are taken from~\cite{DistMultResult}. Other results are taken from the original papers. Best results are written in bold.}
\label{fb15k}
\end{table}

\begin{table*}[!h]
\begin{center}
\resizebox{\textwidth}{!}{
\begin{tabular}{ccccccccccc}
\hline
 &\multicolumn{5}{c}{FB15K-237}&\multicolumn{5}{c}{WN18RR}\cr 
 \cmidrule(lr){2-6}\cmidrule(lr){7-11}
&MR&MRR&Hits@1&Hits@3&Hits@10& MR&MRR&Hits@1&Hits@3&Hits@10\\
\hline
DistMult$^{\diamond}$&254& 0.241 &0.155 &0.263 &0.419 &5110 &0.43 &0.39 &0.44 &0.49\\
ComplEx$^{\diamond}$&339& 0.247& 0.158& 0.275&0.428&5261&0.44&0.41&0.46&0.51\\
ConvE$^{\diamond}$&244&0.325&0.237&0.356&0.501&4187&0.43&0.40&0.44&0.52\\
R-GCN+&-&0.249&0.151&0.264&0.417&-&-&-&-&-\\
RotatE&177&0.338&0.241&0.375&0.533&3340&0.476&0.428&0.492&0.571\\
pRotatE&178&0.328&0.230&0.365&0.524&2923&0.462&0.417&0.479&0.552\\
InteracE&172&0.354&0.263&-&0.535&5202&0.463&0.430&-&0.528\\
QuatE$^2$&-&\textbf{0.366}&0.271&\textbf{0.401}&0.556&-&0.482&0.436&0.499&0.572\\
\hline
GeomE2D&155&0.363&0.269&0.399&0.552&3199&0.483&0.439&0.499&0.571\\
GeomE3D&151&0.364&0.270&0.399&0.555&3303&0.481&0.441&0.494&0.564\\
GeomE+&\textbf{145}&\textbf{0.366}&\textbf{0.272}&\textbf{0.401}&\textbf{0.557}&\textbf{2836}&\textbf{0.485}&\textbf{0.444}&\textbf{0.501}&\textbf{0.573}\\
\hline
\end{tabular}
}
\end{center}
\vspace{-0.1cm}
\caption{Link prediction results on FB15K-237 and WN18RR. $\diamond$ indicates that results are taken from~\cite{ConvE}. Best results are written in bold.}
\vspace{-0.3cm}
\label{fb237}
\end{table*}

\paragraph{Link prediction:}Results on four datasets are shown in Tables~\ref{fb15k} and~\ref{fb237}.  GeomE3D and GeomE2D, as single models, surpass other baselines on FB15K regarding all metrics. GeomE3D and GeomE2D achieve the state-of-the-art results on WN18 except Hits@10 and MR. On FB15K-237 and WN18RR where the local information is less salient, the results of GeomE3D and GeomE2D are close to QuatE$^{2}$. GeomE2D achieves the best MR, MRR and Hits@3 on WN18RR, and GeomE3D achieves the best MR on FB15K-237 as well as the best Hits@1 on WN18RR. 

By combining GeomE2D and GeomE3D, GeomE+ shows more competitive performance. On FB15K, FB15K-237 and WN18RR, GeomE+ outperforms all baselines regarding all metrics. Especially on FB15K, GeomE+ improves MRR by 2.1\%, Hits@1 by 1.7\%, Hits@3 by 2.1\% and Hits@10 by 1.6\%, compared to QuatE$^{2}$. On WN18, GeomE+ achieves the state-of-the-art results regarding MR, MRR, Hits@1 and Hits@3, and achieves the second highest numbers on Hits@10.

\paragraph{The effect of the grade of the multivector space:}Our approach can be generalized in the geometric algebras $\mathbb{G}^{N}$ with different grades $N$. In this paper, we mainly focus on GeomE models embeded in $\mathbb{G}^{2}$ and $\mathbb{G}^{3}$. We do not use multivectors with higher grade $N>3$ in this paper because that would increase the time consumption and memory sizes of training GeomE models and the results of GeomE3D and GeomE2D on the four benchmarks are close. On the other hand, we also test the performances of GeomE1D where each multivector consists of a scalar plus a vector, and find the results drop since the 1-grade multivectors lose some algebra properties after bivectors which square is $-1$ are removed.
\begin{table}[h]
\begin{center}
\resizebox{0.8\textwidth}{!}{
\begin{tabular}{ccccccccc}
\hline
 &\multicolumn{4}{c}{FB15K-237}&\multicolumn{4}{c}{WN18RR}\cr 
 \cmidrule(lr){2-5}\cmidrule(lr){6-9}
&MRR&Hits@1&Hits@3&Hits@10&MRR&Hits@1&Hits@3&Hits@10\\
\hline

GeomE1D&0.355&0.261&0.391&0.545&0.453&0.409&0.465&0.541\\
\hline
\end{tabular}
}
\caption{Link prediction results of GeomE1D on FB15K-237 and WN18RR.}
\end{center}
\vspace{-0.5cm}
\end{table}

\begin{figure}[h!]
\centering
\includegraphics[width=0.9\textwidth]{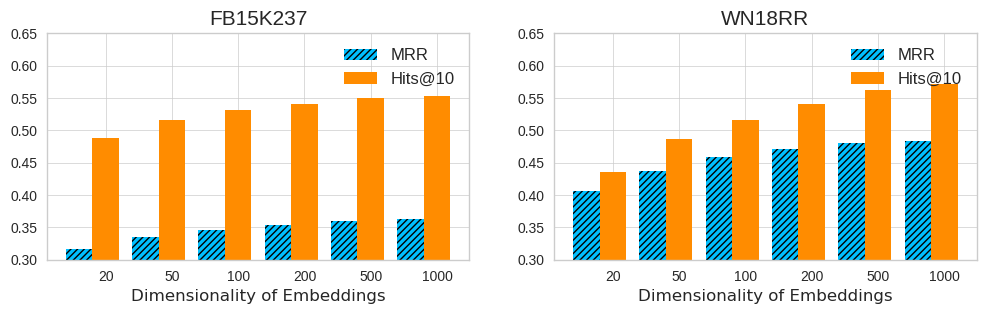}
\vspace{-0.2cm}
\caption{Results of GeomE2D with different dimensionalities on FB15K-237 and WN18RR}
\vspace{-0.5cm}
\label{fig:GeomE2D}
\end{figure}
\paragraph{The effect of the embedding dimensionality:}Figure~\ref{fig:GeomE2D} shows the link prediction results of GeomE2D models with different embedding dimensionalities $k=\{20,50,100,200,500,1000\}$ on FB15K-237 and WN18RR regarding MRR and Hits@10. It can be seen that the performances of GeomE2D improve with the increasing of the embedding dimensionality. We follow the previous work~\cite{QuatE,RotatE} to set the maximum dimensionality to 1000 in order to avoid too much memory and time consumption. It will still be interesting to explore the performances of GeomE models with higher-dimensional embeddings, e.g., Ebisu et al.~\shortcite{toruse} use 10000-dimensional embeddings for TorusE.

\begin{figure}[h]
\centering
\includegraphics[width=0.6\textwidth]{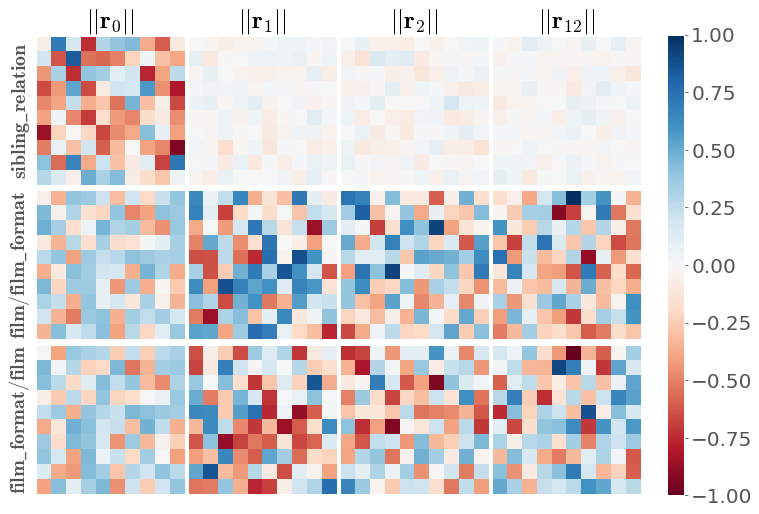} 
\vspace{-0.2cm}
\caption{Visualization of the embeddings of symmetric and inverse relations. 100-dimensional embeddings are reshaped into $10\times 10$ matrices here for a better representation.}
\label{fig:relation}
\vspace{-0.3cm}
\end{figure}
\paragraph{Modeling symmetry and inversion:} In FB15K, \textit{\textbf{sibling\_relationship}} is a typical symmetric relation. By constraining $\phi(h,\textbf{\textit{sibling\_relationship}},t)\approx\phi(t,\textbf{\textit{sibling\_relationship}},h)$ during the training process, we find that the vector and bivector parts of its embedding learned by a 100-dimensional GeomE2D are close to zero as shown in Figure~\ref{fig:relation}. For a pair of inverse relations \textit{\textbf{film/film\_format}} and \textit{\textbf{film\_format/film}} in FB15K, their embeddings are matually conjugate by 
constraining $\phi(h,\textbf{\textit{film/film\_format}},t)\approx\phi(t,\textbf{\textit{film\_format/film}},h)$. These results support our arguments in Section~\ref{relation_modeling} and empirically prove GeomE's ability of modeling symmetric and inverse relations.

\section{Conclusion}
We propose a new gemetric algebra-based approach for KG embedding, GeomE, which utilizes multivector representations to model entities and relations in a KG with the geometric product. Our approach subsumes several state-of-the-art KG embedding models, and takes advantages of the flexibility and representational power of geometric algebras to enhance its generalization capacity, enrich its expressiveness with higher degree of freedom and enable its ability of modeling various relation patterns. Experimental results show that our approach achieves the state-of-the-art results on four well-known benchmarks.


\section*{Acknowledgements}
This work is supported by the CLEOPATRA project (GA no.~812997), the German national funded BmBF project MLwin and the BOOST project.
\bibliographystyle{acl}
\bibliography{coling2020}

\end{document}